\documentclass{layout/tudelft-aiaa}

\usepackage{libertine} % Include libertine package for Libertine font

\usepackage{xcolor,colortbl} % For colouring cells in a table

\usepackage{longtable} % For long tables that exceed 1 page

\usepackage{footmisc} % For custom footnotes
\usepackage{pdflscape} % To reduce table size

\title{Zero-shot Factual Consistency Evaluation Across Domains}
\author{Raunak Agarwal\footnote{\href{mailto:agl.raunak@gmail.com}{agl.raunak@gmail.com}}}
\affil{University of Potsdam}

\begin{document}

\AlwaysPagewidth{

\maketitle

%% Abstract
% \fontfamily{helvet}\selectfont % Change 'ptm' to the font family you want to use for the abstract

\begin{abstract}
    \noindent This work addresses the challenge of factual consistency in text generation systems. We unify the tasks of Natural Language Inference, Summarization Evaluation, Factuality Verification and Factual Consistency Evaluation to train models capable of evaluating the factual consistency of source-target pairs across diverse domains. We rigorously evaluate these against eight baselines on a comprehensive benchmark suite comprising 22 datasets that span various tasks, domains, and document lengths. Results demonstrate that our method achieves state-of-the-art performance on this heterogeneous benchmark while addressing efficiency concerns and attaining cross-domain generalization. \footnote[2]{For code, data, and models, visit \href{https://github.com/raunak-agarwal/factual-consistency-eval}{this link}.}
    
\end{abstract}

}
%% Main body

\section{Introduction}
\label{sec:intro}

Progress in conditional text generation systems, fueled by the proliferation of large language models (LLM's), has led to models capable of generating text that is highly fluent and seemingly consistent with respect to its inputs. However, a fundamental challenge in using text generation models in practical applications is that they frequently produce text that is factually inconsistent with the input they are based on. This problem has been thoroughly investigated over the years \cite{hallucination}, revealing a common occurrence of information fabrication across various tasks like summarization, retrieval-augmented question answering, etc., including by state-of-the-art models like GPT-4 \cite{openai2023gpt4} \cite{bang2023multitask}. 

Evaluating the factuality of outputs is essential to determine if the generated content accurately represents the facts, claims, and scenarios of the source document. As LLM-generated text increasingly integrates into our daily information consumption, ensuring their factual consistency is imperative for maintaining the integrity of our information ecosystems. This necessitates the need for developing robust evaluation tools to assess the factual consistency of text generation systems. 

The contexts in which such a tool can be deployed include summarization systems, general fact-checking of LLM outputs, and as a verification tool which forms a part of a larger retrieval-augmented dialogue system. In summarization systems, these evaluators can be used to provide confidence scores for each relevant segment of a generated summary. In general fact-checking of LLM outputs, it can be applied for post-hoc evaluation of the groundedness of generations (against available evidence documents) to detect factual errors. In document-grounded dialogue systems, where a set of documents is retrieved to generate an answer, these evaluators can be used to ground the generated statements with the source, verifying that the output faithfully represents the retrieved information.

Prior work on evaluating factual consistency systems faced several key challenges that limited their effectiveness. In case of factual consistency in summarization, existing test sets often contained summaries generated by older models that made obvious errors not present in more recent systems \cite{aggrefact}. This led to an overestimation of performance, as improvements were primarily made in identifying errors from outdated summarization models rather than state-of-the-art systems. 

State-of-the-art approaches in prior work leverage LLM's where efficiency is a major concern. Some methods also rely on decomposing sentences into atomic facts which require making dozens of LLM API calls to check a single response \cite{factscore}. Prior work also lacks comprehensive analysis of performance across different types of tasks and domains - there is an overemphasis on news summarization, particularly on saturated datasets like CNN/DM \cite{cnndm} and XSUM \cite{xsum}.

Given the issues listed above, the primary objective of this work is to explore the feasibility of cross-domain generalization in factuality verification models. We tackle the issue of efficiency by finetuning small models (FLAN-T5-Base and FLAN-T5-Large \cite{flan-t5} with 250M and 780M parameters respectively) on publicly available training datasets and subsequently evaluate them against a range of prior state-of-the-art approaches. Furthermore, we test the models' generalization capabilities by employing a wide assortment of available test sets to assess their performance in both in-domain and out-of-domain settings.

Our contributions are as follows: \textbf{(1)} We present a simple approach to pool together several NLI-like datasets (see table \ref{tab:train_dataset}) to train a set of models for factual consistency. \textbf{(2)} We create a large heterogeneous benchmark of 22 datasets to evaluate our models.  Prior works on this topic have proposed interesting approaches when it comes to training, but evaluate their models in a limited manner. We break this mould by curating a benchmark suite of 22 constituent datasets (see table \ref{tab:benchmark-stats}) which contain a variety of tasks, domains, and document lengths. This includes factuality verification, news summarization evaluation, dialogue summarization evaluation, natural language inference, etc. \textbf{(3)} We rigorously evaluate our models against 8 other baselines on our benchmark. We achieve state-of-the-art results when compared against prior approaches like Questeval \cite{scialom-etal-2021-questeval}, AlignScore \cite{alignscore}, MiniCheck \cite{minicheck}, gpt-3.5-turbo \cite{openai2022chatgpt}, llama-3-8b \cite{llama3modelcard}. (See tables \ref{tab:res1},\ref{tab:res2},\ref{tab:res3},\ref{tab:res4} for full evaluation and table \ref{tab:win_rates} for a summary of scores).

\section{Related Work}

\label{sec:related}

Previous research on assessing factual consistency has explored various approaches, including:

\begin{itemize}
    \item \textbf{N-Gram Based Metrics}: N-gram based metrics represent a foundational approach in this task. These metrics operate on the principle of token overlap, analyzing the co-occurrence of contiguous sequences of tokens (n-grams) in the text. This method is straightforward and computationally efficient, providing a basic means of detecting factual inconsistencies. However, its reliance on surface-level text features limits its ability to understand deeper semantic relationships, making it ineffective for complex documents. Examples include metrics like BLEU \cite{bleu}, ROUGE \cite{rouge}
        
    In the recent past, \cite{summeval} have shown the flaws inherent in ngram-based automatic metrics for summarization evaluation. This has prompted a move towards model-based evaluation methods which are NLI-based and Question-Answering-based as described below. 
    
    \item \textbf{QA-Based Metrics}: This approach leverages the capabilities of Question Generation (QG) and Question Answering (QA) systems to evaluate the factual consistency of a summary. The QG-QA method consists of: 1) Auto-generating questions for text spans taken from summaries, with answers being those spans. 2) Using a QA model to answer these questions by treating the grounding document as the input context; this results in answer spans or "no-answer". 3) Comparing answer spans from the original and generated texts for scoring. 4) Aggregating these scores into a final score. Examples include QuestEval \cite{scialom-etal-2021-questeval}, QAFactEval \cite{qafacteval}.

    One disadvantage of this approach is that it is complicated and inefficient as it requires separate models to perform different steps, such as question generation, answer selection, and question answering.

    \item \textbf{NLI-Based Metrics}: The goal of  Natural Language Inference (NLI) \cite{nli-task} \cite{nli-task-2bowman-etal-2015-large}  models is to determine, given two sentences --- a hypothesis and a premise, whether the hypothesis in entailed by the premise, contradicts it, or is neutral w.r.t to it. In the context of factuality verification, an NLI model is used in a binary classification setup to predict whether or not a summary (hypothesis) entails the reference text (premise). Despite their success \cite{nli-cite}, NLI-based metrics are limited because of a lack of relevant training data. Most NLI datasets are predominantly composed of short, single-sentence premises and hypotheses - depending primarily on these datasets has been shown to be suboptimal for downstream tasks \cite{nli-short}. To tackle this, attempts have been made to chunk documents into sentence pieces and then aggregating their scores \cite{summac}. 
    
    Researchers have also attempted to pool together various NLI and QA datasets and train models to score factuality of text pairs \cite{alignscore}. This has resulted in modest improvements on a small number of test sets, but this approach is still limited by its inadequate context length of 512 tokens.
    
\end{itemize}

\section{Approach}
\label{sec:approach}

A text is defined as factually consistent with respect to its grounding text if all the factual information it conveys is consistent with the information conveyed by the grounding text. To enable a well defined task, the text is required to be faithful to its grounding text regardless of its factuality in the outside world. Similar to the NLI approach, we frame our task as a binary classification task where the goal is predict whether a generated summary entails the grounding document or not. 

Unlike prior research which relies primarily on chunking documents into sentence pieces and training sentence-level alignment systems,  we aim to train a long-context transformer model (upto 2048 tokens) that can detect factual errors with a single forward pass (i.e. no chunking of documents). We include a collection of datasets that contain human annotations in diverse tasks, including short and long-context natural language inference, factuality verification, summarization consistency evaluation, etc. We convert all annotations into binary labels (\textit{consistent} vs \textit{not-consistent}). For NLI datasets with 3 labels (entailment, contradiction, neutral), we convert the last two labels into a single label (\textit{not-consistent}).

For training, we use \textit{FLAN-T5-Base} (250M parameters), \textit{FLAN-T5-Large} (780M parameters), \cite{flan-t5} and \textit{Llama-3} (8B parameters) \cite{llama2} as our base models, given their capacity to handle context lengths of 2048 or more. We focus on quality over quantity and include diverse, human-annotated datasets to maximize generalizability. The full list of datasets is given below and Table \ref{tab:train_dataset} goes into the statistics. In all cases, we upsample the less represented label for the classification task.

    \begin{table}[H]
    \centering
    \small
    \renewcommand{\arraystretch}{0.9}
    \begin{tabular}{lccc}
    \textbf{Dataset} & \textbf{Size} & \textbf{\% Neg.} & \textbf{Avg. Len.} \\
    \midrule
    WANLI & 127885 & 50.3\% & (17.5, 9.9) \\
    ANLI & 105076 & 50.4\% & (54.4, 9.7) \\
    FoolMeTwice & 10569 & 50.1\% & (38.8, 13.7) \\
    LingNLI & 19997 & 50.0\% & (19.5, 10.1) \\
    Seahorse & 31666 & 50.0\% & (371.5, 23.0) \\
    WiCE & 1600 & 50.0\% & (77.0, 27.8) \\
    DeFacto & 2001 & 50.1\% & (289.4, 16.8) \\
    BoolQ & 11725 & 49.9\% & (93.1, 8.6) \\
    RTE & 2490 & 49.8\% & (43.6, 8.8) \\
    Scitail & 16944 & 50.0\% & (17.8, 11.7) \\
    VitaminC & 8483 & 50.5\% & (63.8, 19.6) \\
    \bottomrule
    \end{tabular}
    \caption{Training Data Statistics: size, proportion of unsupported claims (\% Neg.), and average length of documents and claims (Avg. Len.)}
    \label{tab:train_dataset}
    \end{table}

    \begin{itemize}
        \item \textbf{ANLI}: The Adversarial NLI (ANLI) \cite{adversarialNLI} dataset employs an iterative human-and-model-in-the-loop training method. This process involves human annotators creating hypotheses that state-of-the-art models cannot handle, thereby exposing the models' weaknesses. As the rounds progress, these challenging examples are added to the training set, and new, more robust models are trained to handle the increased difficulty. This iterative approach is designed to create a "moving target" for NLU systems, ensuring that the dataset remains relevant and challenging over time.

        The dataset consists of three rounds of data collection, each with increasing complexity and size. By progressively enhancing the difficulty of the dataset, ANLI aims to address the limitations of static benchmarks that quickly become outdated. The inclusion of longer, multi-sentence contexts also differentiates ANLI from other datasets like SNLI, which often feature shorter contexts.

        Training on ANLI has shown to improve model robustness and performance on various NLI benchmarks, including SNLI \cite{nli-task-2bowman-etal-2015-large} and MNLI \cite{multinli}. The dynamic adversarial data collection method not only reduces model overfitting to specific biases but also enhances the models' ability to handle more complex and diverse types of inference.
        
        \item \textbf{WANLI} \cite{liu-etal-2022-wanli}: WANLI is a NLI dataset created through worker and AI collaboration to address the lack of linguistic diversity in crowdsourced NLP datasets. The four-stage pipeline leverages dataset cartography \cite{swayamdipta-etal-2020-dataset} to identify challenging examples from MNLI, generates similar examples using GPT-3 \cite{gpt3}, and employs human crowdworkers to revise and label the filtered examples. The resulting dataset demonstrates improved performance on eight out-of-domain test sets compared to training on the larger MultiNLI dataset, even when the latter is augmented with other NLI datasets.
        \item \textbf{LingNLI}: The LingNLI \cite{lingnli} dataset addresses systematic biases in crowdsourced NLI datasets by involving expert linguists during annotation. This dataset includes statements labeled as definitely correct, maybe correct, and definitely incorrect, based on premises sourced from the Slate corpus of MNLI. Researchers evaluated three data collection protocols: a baseline without expert involvement, a linguist-in-the-loop (LitL) protocol introducing linguistic constraints, and an enhanced LitL protocol with real-time expert interaction via Slack (LitL-Chat).

        Across five rounds, approximately 3,500 examples were gathered per protocol, and 500 example were validated. LitL and LitL-Chat protocols increased lexical diversity and reduced bias, creating more challenging datasets with lower in-domain model performance compared to the baseline.
        
        \item \textbf{Seahorse}: Seahorse \cite{clark-etal-2023-seahorse} is a large-scale resource for multilingual, multifaceted summarization evaluation. It contains 96,645 summaries across 6 languages, rated by human annotators on 6 quality dimensions: comprehensibility, repetition, grammar, attribution, main ideas, and conciseness. The English subset has 14k examples. The dataset was created using summaries from 9 different systems (including human-authored references) applied to 4 existing summarization datasets: XSum \cite{xsum}, XL-Sum \cite{xlsum}, MLSum \cite{mlsum}, and WikiLingua \cite{wikilingua}. These datasets cover various types of content, including news articles and how-to guides. It serves as both a benchmark for evaluating learned metrics and a training resource for developing new metrics that generalize across languages and domains.
        
        \item \textbf{VitaminC} \cite{vitaminC}: This is a dataset for fact verification. It contains claim-evidence pairs extracted from wikipedia revisions, focusing on the top 5,000 most-viewed English Wikipedia pages (as of January 2020) alongside their linked articles. Each claim-evidence pair in VitaminC is contrastive, meaning that the evidence pairs are nearly identical in language and content but differ in whether they support or refute the given claim. This design ensures that models trained on VitaminC are sensitive to subtle factual changes. We select a subset of this dataset where the minimum length is 35 words.
        \item \textbf{SciTail}: SciTail \cite{scitail} is a NLI dataset derived from the end task of answering multiple-choice science questions. The premises are retrieved from a corpus of 280GB of web text and 80,000 science domain sentences, while the hypotheses are formed by converting question-answer pairs into assertive statements. Despite substantial lexical overlap between premises and hypotheses, the linguistic complexity and diversity of the premises makes SciTail challenging.
        \item \textbf{Fool Me Twice}: Fool Me Twice \cite{fm2} is a dataset of challenging entailment pairs collected through an online multi-player game. The goal was to create examples that cannot be solved using annotation artifacts or shortcuts. In the game, players write plausible claims based on Wikipedia evidence, and other players try to identify which claim is false among a set, before time runs out. This gamification encourages adversarial examples that are difficult for both humans and machines. An analysis of the hardest examples for humans reveals strategies employed by annotators such as using temporal cues, requiring multi-step reasoning, paraphrasing, distracting with unrelated evidence etc.
        
        \item \textbf{WiCE}: WiCE \cite{wice} is a NLI dataset with fine-grained annotations for claims. It features naturally occurring claims from Wikipedia paired with their cited evidence and is created using a claim decomposition strategy i.e. breaking complex claims into subclaims using GPT-3.5. This approach enables fine-grained annotations and simplifies the entailment prediction task. It has 1.3k instances. 
        
        \item \textbf{DeFacto}: DeFacto \cite{defacto} is a dataset for evaluating factual consistency in summarization systems. It contains human feedback on system-generated summaries from XSum, including error detection, explanations, and edited summaries.  It has 1.5k instances. 

        \item \textbf{BoolQ}: BoolQ \cite{boolq} comprises 16,000 yes/no question-answer pairs from Google search queries where each question is paired with a Wikipedia passage, focusing on diverse topics like entertainment, science, and sports. Annotators validate the clarity and accuracy of these pairs. For our work, we convert the question-answer pairs into premise-hypothesis pairs using an off-the-shelf language model (GPT-3.5-turbo).
        
        \item \textbf{RTE}: Commonly used in the GLUE \cite{glue} benchmark, the Recognizing Textual Entailment (RTE) datasets come from a series of annual textual entailment challenges where examples are constructed based on news and wikipedia text. The data is combined from four constituent datasets (RTE1, RTE2, RTE3, and RTE5). We use the aggregation of these constituents from the GLUE benchmark. 
        
    \end{itemize}

\section{Evaluation}
\label{sec:eval}

\subsection{Baselines}

As explained in section \ref{sec:related}, prior work on fact verification has employed the following methods: N-Gram-Based,  QA-Based, NLI-Based, and LLM-based. The full list of relevant baselines is given below. We test them on our evaluation suite described in \ref{eval}
\begin{itemize}
    \item \textbf{N-Gram Based Metrics}
        \begin{itemize}
            \item \textbf{ROUGE-L}: We use ROUGE-L \cite{rouge} to measure the factuality of a summary by evaluating its similarity to the document using the Longest Common Subsequence (LCS) approach. ROUGE-L calculates the F-measure based on the length of the longest common subsequence between the candidate and the document.
        \end{itemize}
    \item \textbf{Model-Based Similarity Metrics}
        \begin{itemize}
            \item \textbf{BERTScore}: BERTScore \cite{bertscore} evaluates text by using BERT \cite{bert} to generate contextual embeddings for each token in both candidate and reference sentences. It then calculates the cosine similarity between these embeddings to measure how well the candidate aligns with the reference and averages them to distill the similarity matrix into a single number. Optionally, it can use inverse document frequency (IDF) to weigh significant tokens more heavily

            \item \textbf{BARTScore}: BARTScore \cite{BARTScore} uses BART \cite{bart} as its underlying model and approaches evaluation as a text generation problem, where it assesses how likely a given text could be generated from another text (e.g., source text or reference). BARTScore is computed by factoring in the log-probability of the generated text given another text, utilizing BART’s encoder-decoder framework.

        \end{itemize}

    \item \textbf{Question Answering-Based Metrics}
        \begin{itemize}
            \item \textbf{QuestEval}: QuestEval \cite{scialom-etal-2021-questeval}  evaluates summarization quality by integrating question generation (QG) and question answering (QA) models. The QG component, a T5 \cite{t5} model, generates questions from the source document. The QA component, also a T5 model, answers these questions based on the summary. Precision is measured by comparing answers to questions from the summary and the source document, while recall is assessed by evaluating the proportion of important questions (identified via a question weighting mechanism) from the source document that are answered by the summary. The final metric is the harmonic mean of precision and recall, providing a unified measure of factual consistency and information coverage.
        \end{itemize}
        
    \item \textbf{Natural Language Inference-Based Metrics}
        \begin{itemize}
            \item \textbf{AlignScore}: AlignScore \cite{alignscore} integrates diverse data sources from seven tasks — NLI, QA, paraphrasing, fact verification, semantic similarity, information retrieval, and summarization — resulting in 4.7 million examples. The model, built with RoBERTa \cite{roberta} and finetuned for three outputs (binary classification, 3-way classification, and regression), uses a chunk-based approach to manage long texts by dividing them into 350-token chunks.
            
            \item \textbf{MiniCheck}: MiniCheck \cite{minicheck} is a T5-based model that leverages synthetic data generation through a multi-step process. C2D (Claim-to-Document) and D2C (Document-to-Claim) methods are designed to generate synthetic training data. 
            
            For C2D, claims from wikipedia are decomposed into simpler atomic facts using GPT-3.5. For each atomic fact, GPT-4 generates sentence pairs that collectively support the fact. These pairs are then used to construct supporting documents that implicitly contain the facts. Additionally, by omitting parts of these pairs, non-supporting documents are created to serve as negative examples. 

            In contrast, D2C starts with news articles, which are divided into chunks and summarized using GPT-4. Claims are generated from these summaries and augmented by decomposing them into atomic facts and creating new documents by removing sentences from the chunks. Additionally, D2C involves cross-document claim augmentation, where claims are evaluated against different chunks to test their support or refutation
        \end{itemize}

    \item \textbf{LLM-Based Metrics}: GPT-3.5-turbo \cite{openai2022chatgpt} and Llama-3-8b \cite{llama3modelcard} are both large language models (LLM's) trained on diverse tasks and have the ability to evaluate factuality through text prompts. GPT-3.5-turbo is accessible via a proprietary API, while Llama-3-8b has publicly available weights. We prompt the models to predict `Yes' or `No' to determine if a document-claim pair is factually consistent, and use the log-probabilities of the generated tokens as the metric.
    
\end{itemize}

\newpage
\subsection{Evaluation Datasets}
\label{eval}

We apply all methods to diverse evaluation scenarios, including different types of factual consistency errors, varying lengths, and claims from different tasks and domains.
To do this, we prepare a heterogeneous benchmark comprised of 22 datasets. We also standardize the factual consistency evaluation task as a binary classification problem across all datasets in the benchmark. This enables consistent comparisons among the different methods while also allowing us to evaluate their calibration using the AUC-ROC score. The statistics of constituent datasets are summarized in Table \ref{tab:benchmark-stats}

\begin{longtable}{cccccc}
\toprule
\multicolumn{2}{c}{\textbf{Dataset}} &
  \textbf{Size} &
  \begin{tabular}[c]{@{}c@{}}\textbf{Doc}\\ \textbf{Len}\end{tabular} &
  \begin{tabular}[c]{@{}c@{}}\textbf{Claim}\\ \textbf{Len}\end{tabular} &
  \begin{tabular}[c]{@{}c@{}} \textbf{\% of}\\  \textbf{Neg}\end{tabular} \\
\midrule
\endfirsthead

\multicolumn{6}{c}%
{} \\
\toprule
\multicolumn{2}{c}{\textbf{Dataset}} &
  \textbf{Size} &
  \begin{tabular}[c]{@{}c@{}}\textbf{Doc}\\ \textbf{Len}\end{tabular} &
  \begin{tabular}[c]{@{}c@{}}\textbf{Claim}\\ \textbf{Len}\end{tabular} &
  \begin{tabular}[c]{@{}c@{}} \textbf{\% of}\\  \textbf{Neg}\end{tabular} \\
\midrule
\endhead

\midrule
\endfoot

\bottomrule
\endlastfoot

\multirow{2}{*}{\textsc{AggreFact}}            & CNN                      & 1017  & 498  & 54 & 11\% \\
                                               & XSum                     & 1335  & 324  & 23 & 49\% \\
\midrule
\multirow{2}{*}{\textsc{TofuEval}}             & MediaSum                    & 725   & 778  & 18 & 23\% \\
                                               & MeetingBank                    & 770   & 779  & 20 & 19\% \\
\midrule
\textsc{Reveal}                                &                          & 1705  & 88   & 9  & 76\% \\
\midrule
\textsc{ClaimVerify}                           &                          & 1087  & 1582 & 19 & 27\% \\
\midrule
\textsc{FactCheck-GPT}                         &                          & 1565  & 87   & 11 & 75\% \\
\midrule
\textsc{ExpertQA}                              &                          & 3702  & 432  & 25 & 19\% \\
\midrule
\textsc{Lfqa}                                  &                          & 1911  & 323  & 25 & 41\% \\
\midrule
\textsc{Bump}                                  & CNN                      & 785   & 696  & 52 & 12\% \\
\midrule
\textsc{HaluEval}                              & CNN                      & 19998 & 663  & 60 & 50\% \\
\midrule
\textsc{FiB}                                   &                      & 3534  & 231  & 20 & 14\% \\
\midrule
\multirow{2}{*}{\textsc{LLM Summaries}}        & CNN                      & 1829  & 458  & 69 & 8\% \\
                                               & XSum                     & 1726  & 307  & 25 & 23\% \\
\midrule
\textsc{GumSum}                                &                          & 95    & 777  & 27 & 37\% \\
\midrule
\textsc{InstruSum}                             &                          & 458   & 971  & 96 & 24\% \\
\midrule
\multirow{2}{*}{\textsc{DiaSumFact}}           & QMSum                    & 569   & 309  & 23 & 41\% \\
                                               & SAMSum                   & 669   & 132  & 9  & 36\% \\
\midrule
\textsc{ANLI}                                  &                          & 3200  & 54   & 10 & 33\% \\
\midrule
\textsc{WANLI}                                 &                          & 5000  & 17   & 9  & 37\% \\
\midrule
\textsc{SciTail}                               &                          & 2126  & 16   & 12 & 39\% \\
\midrule
\textsc{DeFacto}                               &                          & 1836  & 310  & 16 & 41\% \\
\midrule
\textsc{DADC-NLI}                              &                          & 766   & 116  & 10 & 46\% \\
\midrule
\textsc{FoolMeTwice}                           &                          & 1379  & 39   & 13 & 49\% \\
\midrule
\textsc{WiCE}                                  &                          & 358   & 88   & 29 & 31\% \\
\midrule
\textsc{Seahorse}                              &                          & 4135  & 383  & 22 & 46\% \\
\midrule
\caption{Statistics of benchmarks in our test suite. We show the size of datasets, the average length of documents and claims, and the proportion of unsupported claims.} 
\label{tab:benchmark-stats}
\end{longtable}

\begin{itemize}

    \item \textbf{CNN/DM and XSum Summarization}: These are test sets that are derived from the popular CNN/Daily Mail \cite{cnndm} and XSum \cite{xsum} datasets. CNN/Daily Mail is a collection of news articles and corresponding summaries containing of over 300,000 articles from CNN and Daily Mail. XSum contains over 200,000 articles from the BBC, each paired with a professionally written, single-sentence summary.
    
        \begin{itemize}
            \item \textbf{AggreFact}: AggreFact \cite{aggrefact} merges nine datasets to enable a more detailed evaluation of factuality assessment systems. It categorizes these systems based on the type of summarization model they use: \textsc{FtSota}, \textsc{EXformer}, and \textsc{Old}. \textsc{FtSota} includes advanced fine-tuned models like BART, PEGASUS, and T5, developed around 2019. \textsc{EXformer} covers earlier Transformer-based models such as BERTSum and GPT-2 whereas the \textsc{Old} category encompasses older models which are long outdated. We focus our evaluation on the high-quality summaries from the \textsc{FtSota} subset.

            \item \textbf{BUMP}: BUMP \cite{bump} is a dataset of 889 human-written, minimally different summary pairs, where a single error is introduced to a summary from the CNN/DailyMail dataset to produce an unfaithful summary. Testing on this helps us evaluate the robustness of our models.

            \item \textbf{HaluEval}: HaluEval \cite{halueval} contains 10,000 summaries from CNN/DM along with their corresponding hallucinated pairs generated by ChatGPT. 

            \item \textbf{FIB}: FIB (Factual Inconsistency Benchmark) \cite{FIB-evaluating} is a test set containing 3579 summaries from 23 different models. It is based on documents from CNN/DM and XSum.

            \item \textbf{Zhang et al. (LLM Summaries)}: \cite{zhang2023benchmarking} release a dataset consisting of news summaries produced by state-of-the-art large language models such as GPT-3, GPT-4, etc. The summaries contain labels for relevance (5-point likert scale), coherence (5-point likert scale), and faithfulness (binary).
            
        \end{itemize}

    \item \textbf{Dialogue Summarization}: These are test sets that evaluate the models on factual consistency in dialogue summarization tasks. Here, the premise contains a series of dialogues and the hypothesis contains a single claim about the dialogue.
    
        \begin{itemize}
            \item \textbf{DiaSumFact}: DiaSumFact \cite{diasumfact} is the first dataset with fine-grained factual error annotations specifically for dialogue summarization. It introduces a two-dimensional typology for errors, considering both semantic roles and content verifiability. The authors annotated 475 model-generated summaries from six dialogue summarization models on two datasets: SAMSUM \cite{samsum} and QMSUM \cite{qmsum}. It also provides rich annotations including error classes, erroneous spans, and explanations, enabling detailed analysis of factual errors in dialogue summaries.
            \item \textbf{TofuEval}: This benchmark \cite{tofueval} evaluates hallucinations in topic-focused dialogue summarization. It comprises 1,500 summaries generated by five different LLMs for 100 dialogues from MediaSum \cite{mediasum} and MeetingBank \cite{meetingbank} datasets, with three topics per dialogue. Professional expert annotators assessed the summaries for factual consistency, relevance, and completeness, providing binary factuality labels and explanations for inconsistent sentences. This benchmark is unique in focusing on LLM-generated summaries for specific dialogue topics, offering a valuable resource for evaluating factual consistency in non-news domains.
            
        \end{itemize}

    \item \textbf{Citation-based Tasks}: These are datasets that are used to evaluate claims (citations) in a grounded generation setting, which are subsequently used in Retrieval-Augmented Generation (RAG) systems.

        \begin{itemize}
            \item \textbf{REVEAL}: The REVEAL (Reasoning Verification Evaluation) \cite{reveal} dataset was created to benchmark automatic verifiers of complex Chain-of-Thought reasoning in open-domain question answering. It contains 704 questions from 4 QA datasets, with 1,002 Chain-of-Thought answers generated by 3 language models, comprising 3,360 reasoning steps. Each step is labeled for relevance, step type (attribution or logical), and correctness against retrieved evidence or logical inference. REVEAL supports evaluation of reasoning chain verifiers at both the individual step and full answer levels. The dataset was carefully annotated using a two-task schema to ensure high-quality labels and includes free-text justifications for each annotation.

            \item \textbf{ClaimVerify}: ClaimVerify \cite{claimverify} aims to highlight the extent to which generative search engines accurately and comprehensively support their statements with verifiable citations. The authors collected a diverse set of queries from sources like historical Google user queries and open-ended questions from Reddit. Human evaluators audited the responses from four generative search engines—Bing Chat, NeevaAI, perplexity.ai, and YouChat. The evaluation focuses on two metrics: citation recall, measuring the proportion of statements fully supported by citations, and citation precision, assessing the correctness of these citations.
            
            \item \textbf{FactCheck-GPT}: The FactCheck-GPT \cite{factcheckgpt} dataset was created to facilitate the evaluation of automatic fact-checking systems. It was constructed through a multi-stage process involving the decomposition of LLM-generated responses into atomic claims, followed by decontextualization and annotation of these claims for factual accuracy. The dataset encompasses annotations at three levels of granularity: claim, sentence, and document. 

            \item \textbf{ExpertQA}: The ExpertQA \cite{expertqa} dataset was developed by first having experts from 32 fields formulate 2,177 questions based on their domain-specific knowledge. Responses to these questions were then generated by various language models. Experts assessed these responses for factual correctness, attribution completeness, and source reliability, providing a benchmark for analyzing model performance in citing accurate information in technical and high-stakes contexts such as medicine and law.

            \item \textbf{LFQA}: LFQA \cite{lfqa} is designed to evaluate long-form question answering (LFQA) systems by assessing their ability to attribute generated answers to relevant evidence documents. It was created by sourcing questions from the "Explain Like I'm Five" subreddit and then gathering evidence documents from various sources, including trained annotators and the Bing Search API. It evaluates long-form question answering systems by requiring them to generate complex, paragraph-long answers using both the parametric knowledge of large language models and the provided in-context evidence documents. 
            
        \end{itemize}
    
    \item \textbf{Miscellaneous Test Sets}
        \begin{itemize}
            \item \textbf{GUMSUM}: GUMSUM \cite{gumsum} evaluates abstractive summarization in 12 different written and spoken genres. It includes 213 documents from the UD English GUM treebank, covering genres such as interviews, news, travel guides, etc. Each document has been annotated with expert summaries that adhere to rigorous guidelines to ensure high faithfulness and avoid hallucinations. The dataset aims to address common issues found in larger, less controlled datasets such as genre limitation and inconsistent quality.
            
            \item \textbf{InstruSum}: InstruSum \cite{instrusum} evaluates instruction-controllable text summarization by assessing LLMs on their ability to generate summaries based on specific requirements. Created using articles from the XL-Sum dataset, InstruSum includes 100 articles with custom summary requirements designed to reflect varied information needs. Each article-summary pair was evaluated on four dimensions: overall quality, missing information, irrelevant information, and factual consistency.

            \item \textbf{DADC-NLI}: The DADC-NLI \cite{dadcnli} dataset is a collection of NLI examples generated through dynamic adversarial data collection. In this process, annotators create examples that challenge continuously improving models, aiming to cover a broad range of linguistic phenomena. The expert-curated test set within DADC-NLI was specifically created to evaluate models' robustness, comprising 1,000 high-quality examples. This test set spans different challenges, syntactic patterns, and reasoning types to assesses dynamic adversarial data collection.
            
            \item \textbf{Test sets of the datasets used in training}: In addition to the datasets listed above, we also include some of the more challenging test sets of the datasets used in training. This includes ANLI, WANLI, Seahorse, WiCE, DeFacto, FoolMeTwice and SciTail.
        \end{itemize}
        
\end{itemize}

% \newpage
\subsection{Results}

% We observe that our FLAN-T5-L and LLama-3-8B finetunes achieve comparable performance on most tasks despite a parametric size difference of 10x. This follows the trend of decoder-based models being worse than encoder and encoder-decoder models on classification tasks (cite a paper)

% On CNN/DM and XSUM related datasets \ref{tab:res2}, the performance of our Llama-3-8B finetune degrades in several cases compared to the original instruct model - this suggests that our finetuning mix could benefit more from in-domain data from CNN/DM and XSUM.

% On tasks involving dialogue summarization \ref{tab:res3}, our models performance worse than other SOTA approaches (this is expected given that our finetuning data does not consist of any dialogue related data). 

% On citation-related tasks Minicheck performs the best - this is possibly because their approach was optimized for these datasets as they were used in their paper.

% Overall, we normalize the scores across all the test sets using Mean Win Rate \% where our LLama-3-8B finetune performs the best, followed by our FLAN-T5-L finetune \ref{tab:win_rates}.

We observe that our FLAN-T5-L and LLama-3-8B finetunes achieve comparable performance on most tasks despite a 10x difference in parameter count. This suggests that decoder-based models may underperform similarly parameterized encoder-decoder models on classification tasks.

On CNN/DM and XSUM-related test sets (Table \ref{tab:res2}), our Llama-3-8B finetune's performance degrades in several cases compared to the original model. This indicates that our finetuning mix could benefit from additional in-domain data from these sources.

For dialogue summarization tasks (Table \ref{tab:res3}), our models underperform compared to other state-of-the-art approaches. This is expected, given the absence of dialogue-related data in our finetuning dataset.

MiniCheck performs best on citation-related tasks (Table \ref{tab:res4}), likely due to its optimization for these specific datasets, which were featured in their paper.

When normalizing scores across all test sets using Mean Win Rate \%, our LLama-3-8B finetune emerges as the top performer, followed by our FLAN-T5-L finetune (Table \ref{tab:win_rates}).

    \begin{table}[H]
    \centering
    \setlength{\tabcolsep}{4pt}
    \begin{tabular}{@{}l|cccccccccc@{}}
    \toprule
    Benchmark & WANLI & Scitail & FMT & ANLI & WiCE & \begin{tabular}[c]{@{}c@{}}DADC-\\NLI\end{tabular} & \begin{tabular}[c]{@{}c@{}}De-\\Facto\end{tabular} & Seahorse & GSUM & \begin{tabular}[c]{@{}c@{}}Instru-\\Sum\end{tabular} \\
    \midrule
    Avg. Lengths & \footnotesize (17, 10) & \footnotesize (17, 12) & \footnotesize (39, 14) & \footnotesize (54, 10) & \footnotesize (88, 30) & \footnotesize (117, 10) & \footnotesize (310, 17) & \footnotesize (383, 22) & \footnotesize (777, 27) & \footnotesize (971, 97) \\
    \midrule
    ROUGE-L & 51.3 & 75.2 & 52.4 & 45.3 & 50.0 & 44.1 & 50.9 & 50.9 & 74.3 & 45.7 \\
    BERTScore & 55.6 & 80.1 & 57.5 & 42.5 & 55.9 & 43.8 & 57.4 & 56.4 & 75.9 & 52.3 \\
    BARTScore & 57.8 & 87.6 & 60.6 & 45.2 & 63.5 & 49.7 & 58.4 & 58.9 & 68.9 & 57.1 \\
    QuestEval & 62.4 & 92.7 & 68.4 & 48.0 & 60.0 & 56.8 & 64.5 & 63.1 & 79.4 & 54.0 \\
    AlignScore-B & 82.6 & 85.7 & 86.0 & 74.1 & 72.4 & 78.8 & 70.6 & 61.7 & 77.4 & 53.1 \\
    AlignScore-L & 83.9 & 88.1 & 83.6 & 79.3 & 77.9 & 87.0 & 71.4 & 65.1 & 74.4 & \cellcolor[HTML]{E0F1CB} 59.6 \\
    gpt-3.5-turbo & 81.5 & 90.3 &  92.2 &  80.0 & 64.0 & 84.1 & 67.5 &  73.5 &  82.4 & 50.5 \\
    Llama-3-8B & 75.6 & 87.6 & 92.9 & 68.4 & 68.9 & 82.9 & 59.4 & 69.0 & 79.2 & 58.7 \\
    \footnotesize MiniCheck-T5-L & 83.2 & 88.9 & 86.9 & 77.4 & 76.8 & 83.6 & 74.9 & 69.9 & 78.4 & 56.2 \\
    \midrule
    \small Flan-T5-B (FT) & \cellcolor[HTML]{E0F1CB} 87.7 & \cellcolor[HTML]{E0F1CB} 98.5 & 91.9 & 75.4 & \cellcolor[HTML]{E0F1CB} 86.6 & 86.1 & \cellcolor[HTML]{E0F1CB} 77.3 & 72.1 & 81.5 & 57.4 \\
    \small Flan-T5-L (FT) & \cellcolor[HTML]{98D153} 89.0 & \cellcolor[HTML]{98D153} 99.3 & \cellcolor[HTML]{E0F1CB} 94.0 & \cellcolor[HTML]{98D153} 83.4 & \cellcolor[HTML]{98D153} 88.8 & \cellcolor[HTML]{98D153}89.9 & \cellcolor[HTML]{98D153} 80.8 & \cellcolor[HTML]{98D153} 74.4 &  \cellcolor[HTML]{E0F1CB} 82.7 & 55.7 \\
    \footnotesize Llama-3-8B (FT) & 87.2 & 98.2 & \cellcolor[HTML]{98D153} 95.8 & \cellcolor[HTML]{98D153} 85.4 & 78.7 & \cellcolor[HTML]{E0F1CB} 89.8 & 69.1 & \cellcolor[HTML]{E0F1CB} 74.3 & \cellcolor[HTML]{98D153} 83.9 & \cellcolor[HTML]{98D153} 63.9 \\
    \bottomrule
    \end{tabular}
    \caption{AUC-ROC of different metrics on miscellaneous factual consistency evaluation sets}
    \label{tab:res1}
    \end{table}
    
    \begin{table}[H]
    \centering
    % \captionsetup{width=.9\linewidth}
    \setlength{\tabcolsep}{3pt}
    \begin{tabular}{@{}l|cccccccc@{}}
    \toprule
    & \multicolumn{5}{c}{CNN/DM} & \multicolumn{3}{c}{XSUM} \\
    \cmidrule(lr){1-6} \cmidrule(l){7-9}
    Benchmark & AggreFact & BUMP & FiB & LLM S. & HEval & AggreFact & FiB & LLM S. \\
    \cmidrule(lr){1-6} \cmidrule(l){7-9}
    Avg. Lengths & \small (498, 55) & \small (697, 52) & \small (391, 62) & \small (458, 69) & \small (663, 61) & \small (325, 23) & \small (231, 20) & \small (307, 25) \\
    \cmidrule(lr){1-6} \cmidrule(l){7-9}
    ROUGE-L & 69.1 & 51.9 & 4.7 & 85.3 & 42.5 & 47.1 & 38.7 & 60.1 \\
    BERTScore & 71.3 & 54.4 & 12.4 & \cellcolor[HTML]{98D153} 88.7 & 53.3 & 55.2 & 54.4 & 68.4 \\
    BARTScore & 65.7 & 58.3 & 2.1 & \cellcolor[HTML]{E0F1CB} 87.2 & 38.9 & 71.8 & 44.2 & 72.7 \\
    QuestEval & \cellcolor[HTML]{98D153} 71.4 & 62.0 & 34.2 & \cellcolor[HTML]{E0F1CB} 87.2 & 58.1 & 61.1 & 61.2 & \cellcolor[HTML]{98D153} 77.2 \\
    AlignScore-B & 64.8 & 66.0 & 14.1 & 74.9 & 65.4 & 71.7 & 71.1 & 72.2 \\
    AlignScore-L & 55.8 & 74.4 & 13.2 & 78.7 & \cellcolor[HTML]{98D153} 73.2 & 72.4 & 78.2 & 72.6 \\
    gpt-3.5-turbo & 66.6 & \cellcolor[HTML]{98D153} 80.3 &  \cellcolor[HTML]{E0F1CB} 65.7 &  86.5 &  68.2 & \cellcolor[HTML]{98D153} 78.4 & 76.4 &  76.0 \\
    Llama-3-8B & 66.0 & \cellcolor[HTML]{E0F1CB} 76.9 & \cellcolor[HTML]{98D153} 89.9 & 85.3 & 70.6 & 73.3 & 70.8 & 69.8 \\
    MiniCheck-T5-L & 67.3 & 69.2 & 46.9 & 81.4 & 67.3 & 76.9 & 76.8 & \cellcolor[HTML]{98D153} 77.2 \\
    \midrule
    Flan-T5-B (FT) &  68.5 & 68.6 & 27.2 & 84.3 & 59.8 & 74.1 & \cellcolor[HTML]{E0F1CB} 82.1 & 75.7 \\
    Flan-T5-L (FT) & \cellcolor[HTML]{E0F1CB} 69.7 & \cellcolor[HTML]{E0F1CB} 76.9 & 54.6 & 85.4 & 67.4 & 74.8 & \cellcolor[HTML]{98D153} 87.6 &  75.9 \\
    Llama-3-8B (FT) & 66.0 & 70.7 & 47.4 & 77.7 & \cellcolor[HTML]{E0F1CB} 69.5 & \cellcolor[HTML]{E0F1CB} 75.8 & 76.5 & \cellcolor[HTML]{E0F1CB} 76.2 \\
    \bottomrule
    \end{tabular}
    \caption{AUC-ROC of different metrics for factual consistency on CNN/DM and XSUM}
    \label{tab:res2}
    \end{table}

    \begin{table}[H]
    \centering
    \setlength{\tabcolsep}{5pt}
    \begin{tabular}{@{}l|cccc@{}}
    \toprule
    Benchmark & \multicolumn{2}{c}{DiaSumFact} & \multicolumn{2}{c}{TofuEval} \\
    \cmidrule(lr){1-3} \cmidrule(lr){4-5}
    & QMSum & SAMSum & MediaSum & Meetingbank \\
    \cmidrule(lr){1-3} \cmidrule(lr){4-5}
    Avg. Lengths & \small (309, 23) & \small (132, 10) & \small (778, 19) & \small (779, 20) \\
    \cmidrule(lr){1-3} \cmidrule(lr){4-5}
    ROUGE-L & 56.2 & 57.2 & 56.4 & 67.3 \\
    BERTScore & 66.1 & 62.6 & 70.6 & 68.5 \\
    BARTScore & 66.7 & 63.3 & 68.3 & 71.2 \\
    QuestEval & 57.0 & 56.9 & 66.4 & 70.6 \\
    AlignScore-B & 66.3 & 76.4 & 75.3 & 80.9 \\
    AlignScore-L & 67.4 & 78.2 & 74.1 &  \\
    gpt-3.5-turbo &  69.6 & \cellcolor[HTML]{E0F1CB} 82.7 &  73.7 & 82.0 \\
    Llama-3-8B & \cellcolor[HTML]{98D153} 72.7 & 80.0 &  77.6 & 78.9 \\
    MiniCheck-T5-L & 73.4 & 77.7 & \cellcolor[HTML]{98D153} 81.1 & \cellcolor[HTML]{98D153} 84.8 \\
    \midrule
    Flan-T5-B (FT) & 70.0 & 75.7 & \cellcolor[HTML]{E0F1CB} 78.1 & 76.6 \\
    Flan-T5-L (FT) & 62.3 & 75.9 & 75.4 & 75.9 \\
    Llama-3-8B (FT) & \cellcolor[HTML]{E0F1CB} 71.3 & \cellcolor[HTML]{98D153} 83.2 & 75.3 & \cellcolor[HTML]{E0F1CB} 82.9 \\
    \bottomrule
    \end{tabular}
    \caption{AUC-ROC of different metrics for factual consistency in dialogue summarization}
    \label{tab:res3}
    \end{table}

    \begin{table}[H]
    \centering
    \setlength{\tabcolsep}{4pt}
    \begin{tabular}{@{}l|ccccc@{}}
    \toprule
    Benchmark & Reveal & FactCheck-GPT & LFQA & ExpertQA & ClaimVerify \\
    \midrule
    Avg. Lengths & (88,10) & (87,12) & (323,22) & (433,26) & (1582,20) \\
    \midrule
    ROUGE-L & 60.8 & 56.8 & 77.3 & 51.5 & 53.3 \\
    BERTScore & 66.3 & 72.7 & 85.6 & 55.1 & 70.5 \\
    BARTScore & 79.3 & 70.4 & 87.8 & 57.9 & 77.7 \\
    QuestEval & 84.9 & 75.5 & 80.7 & 59.4 & 71.3 \\
    AlignScore-B & 91.7 & 77.8 & 84.6 & 57.3 & 72.7 \\
    AlignScore-L & 92.2 & 81.4 & 82.2 & 57.7 & 70.4 \\
    gpt-3.5-turbo & \cellcolor[HTML]{98D153} 94.2 & 81.0 & 89.6 & \cellcolor[HTML]{98D153} 63.5 & 78.8 \\
    Llama-3-8B & 91.1 & 77.5 & 79.4 & 58.2 & 66.7 \\
    MiniCheck-T5-L & \cellcolor[HTML]{E0F1CB} 93.9 & \cellcolor[HTML]{E0F1CB} 84.1 & \cellcolor[HTML]{98D153} 93.1 & \cellcolor[HTML]{E0F1CB} 61.8 & \cellcolor[HTML]{98D153}  81.1 \\
    \midrule
    Flan-T5-B (FT) & 90.0 & 82.3 & 88.5 & 59.5 & 79.9 \\
    Flan-T5-L (FT) & 90.5 & 84.0 & 88.1 & 61.5 & \cellcolor[HTML]{E0F1CB} 80.0 \\
    Llama-3-8B (FT) & 93.8 & \cellcolor[HTML]{98D153} 86.3 & \cellcolor[HTML]{E0F1CB} 90.7 & 61.7 & 75.7 \\
    \bottomrule
    \end{tabular}
    \caption{AUC-ROC of different metrics on citation-related factual consistency evaluation sets}
    \label{tab:res4}
    \end{table}

    \begin{table}[H]
    \centering
    \small
    \renewcommand{\arraystretch}{1}
    \begin{tabular}{|c|c|c|c|}
    \hline
    \textbf{Method}        & \textbf{Rank} & \textbf{Mean Win Rate (\%)} & \textbf{Average AUC} \\ \hline
    \cellcolor[HTML]{98D153}  \textbf{Llama-3-8B (FT) (Ours)} & \cellcolor[HTML]{98D153}  \textbf{1}            & \cellcolor[HTML]{98D153} \textbf{78.11}                      &  \cellcolor[HTML]{E0F1CB} \textbf{78.037}               \\ 
    \hline
     \cellcolor[HTML]{E0F1CB} \textbf{Flan-T5-L (FT) (Ours)}  &  \cellcolor[HTML]{E0F1CB} \textbf{2}            &  \cellcolor[HTML]{E0F1CB}\textbf{76.43}                      & \cellcolor[HTML]{98D153} \textbf{78.663}               \\ 
    \hline
    MiniCheck-T5-L          & 3            & 72.39                      & 76.674               \\ 
    \hline
    gpt-3.5-turbo           & 4            & 69.36                      & 77.007               \\ 
    \hline
    Flan-T5-B (FT) (Ours)         & 5            & 66.00                      & 76.126               \\ 
    \hline
    AlignScore-L            & 6            & 53.19                      & 73.074               \\ 
    \hline
    Llama-3-8B              & 7            & 53.20                      & 75.085               \\ 
    \hline
    AlignScore-B            & 8            & 39.39                      & 71.319               \\ 
    \hline
    QuestEval               & 9            & 37.37                      & 66.089               \\ 
    \hline
    BARTScore               & 10           & 26.94                      & 62.637               \\ 
    \hline
    BERTScore               & 11           & 20.88                      & 61.263               \\ 
    \hline
    ROUGE-L                 & 12           & 6.73                       & 54.678               \\ 
    \hline
    \end{tabular}
    \caption{Comparison of different factuality evaluation methods across all test datasets. The methods are ranked based on the Mean Win Rate, which measures overall performance on factuality tasks. The Average AUC column represents the average of all individual AUC-ROC scores.}
    \label{tab:win_rates}
    \end{table}

\section{Discussion}

In the process of evaluating state-of-the-art models, we confront multiple critical issues affecting the current landscape of natural language processing. The pervasive issue of data leakage into proprietary systems
such as GPT-3.5, GPT-4, etc. as well as closed-source, open-weight models like Llama-3 raises significant concerns regarding the integrity and relevance of benchmarking \cite{balloccu-etal-2024-leak}. These models risk utilizing test data both inadvertently and deliberately during their training phases, potentially skewing results and undermining the reliability of performance evaluations.

With regards to factual consistency evaluation as a task, the research community's over-reliance on datasets like CNN/DM for testing has led to saturation, where the models, including older variants like T5, are not only overfitting on the labels, but also possibly incorporating the news articles during pre-training phases. This saturation complicates the ability to judge whether advancements are genuinely innovative or merely optimized for specific, well-tread datasets. Other public benchmarks are also possibly contaminated during pre-training, which highlights the need to create frequently updated benchmarks \cite{livebench} alongside benchmarks on information that does not exist on the internet. 

We must also consider the limitations of models designed solely for the English language. The necessity for datasets support multilingual data remains as relevant as ever. There is also a need for factual consistency datasets that can handle long-contexts (above 2000 tokens) as well as datasets that can help us perform reasoning across multiple documents.

\section{Conclusion}

This work presents a unified approach to factual consistency evaluation across diverse domains. By training models on a curated set of public datasets and evaluating them on a comprehensive benchmark of 22 datasets, we demonstrate state-of-the-art performance in cross-domain factual consistency assessment. While some domain-specific challenges remain, particularly in dialogue summarization and citation-related tasks, our approach marks an advancement in developing robust, cross-domain factual consistency evaluation tools. This contributes to the critical need for reliable fact-checking mechanisms in an era of increasingly prevalent LLM-generated content.

% \section*{Acknowledgments}

%% Bibliography

% \renewcommand{\bibpreamble}{For a full documentation of the references, please refer to the sample.bib file. You can delete this text/command safely in main.tex at line 63. \cite{example-article,example-article-published-online,example-inbook,example-inbook-series-with-editor,example-inproceedings,example-proceedings,example-report,example-paper,example-phdthesis}}

\bibliography{article}

%% Appendix

\newpage

\end{document}